%% file: 0.main.tex
\documentclass[11pt,a4paper]{article}
\usepackage[hyperref]{emnlp2020}
\usepackage{times}
\usepackage{latexsym}

\usepackage{times}
\usepackage{epsfig}
\usepackage{graphicx}
\usepackage{amsmath}
\usepackage{amssymb}
\usepackage{enumitem}

\usepackage{longtable}

\makeatletter
\def\thickhline{%
  \noalign{\ifnum0=`}\fi\hrule \@height \thickarrayrulewidth \futurelet
   \reserved@a\@xthickhline}
\def\@xthickhline{\ifx\reserved@a\thickhline
               \vskip\doublerulesep
               \vskip-\thickarrayrulewidth
             \fi
      \ifnum0=`{\fi}}
\makeatother

\newlength{\thickarrayrulewidth}
\usepackage{array}
\newcolumntype{P}[1]{>{\centering\arraybackslash}p{#1}}

\usepackage{multirow}
\usepackage{booktabs}

\usepackage{pifont}% http://ctan.org/pkg/pifont
\newcommand{\xdownarrow}[1]{%
  {\left\downarrow\vbox to #1{}\right.\kern-\nulldelimiterspace}
}

\usepackage{xcolor}
\definecolor{bittersweet}{rgb}{1.0, 0.44, 0.37}
\definecolor{mygreen}{rgb}{0.29, 0.7, 0.48}

\newcommand\dsetname{\textsc{VLEP}}

\usepackage{colortbl}
\definecolor{Gray}{gray}{0.85}

\usepackage{bbold}

\usepackage{microtype}

\aclfinalcopy % Uncomment this line for the final submission
\title{What is More Likely to Happen Next? \\ Video-and-Language Future Event Prediction}

\author{
  Jie Lei $\;\;\;\;\;$ Licheng Yu $\;\;\;\;\;$ 
  Tamara L. Berg $\;\;\;\;\;$ Mohit Bansal \\
  Department of Computer Science \\ University of North Carolina at Chapel Hill \\
  {\tt \{jielei, licheng, tlberg, mbansal\}@cs.unc.edu} \\
}

\begin{document}
\maketitle
\begin{abstract}
Given a video with aligned dialogue, people can often infer \textit{what is more likely to happen next}. Making such predictions requires not only a deep understanding of the rich dynamics underlying the video and dialogue, but also a significant amount of commonsense knowledge.
In this work, we explore whether AI models are able to learn to make such multimodal commonsense next-event predictions.
To support research in this direction, we collect a new dataset, named Video-and-Language Event Prediction (\dsetname), with 28,726 future event prediction examples (along with their rationales) from 10,234 diverse TV Show and YouTube Lifestyle Vlog video clips. 
In order to promote the collection of non-trivial challenging examples, we employ an adversarial human-and-model-in-the-loop data collection procedure.
We also present a strong baseline incorporating information from video, dialogue, and commonsense knowledge.
Experiments show that each type of information is useful for this challenging task, and that compared to the high human performance on \dsetname, our model provides a good starting point but leaves large room for future work.\footnote{Dataset, code are available at \url{https://github.com/jayleicn/VideoLanguageFuturePred}}

\end{abstract}

%%%%%%%%% BODY TEXT
\input{1.introduction.tex}

\input{2.related_work.tex}
\input{3.dataset.tex}

\input{4.method.tex}

\input{5.experiments.tex}
\input{6.conclusion.tex}

\bibliographystyle{acl_natbib}
\bibliography{emnlp2020}

\appendix
\input{7.appendix}

\end{document}

%% file: 1.introduction.tex
\section{Introduction}\label{sec:intro}
\begin{figure}[t]
    \centering
  \includegraphics[width=\linewidth]{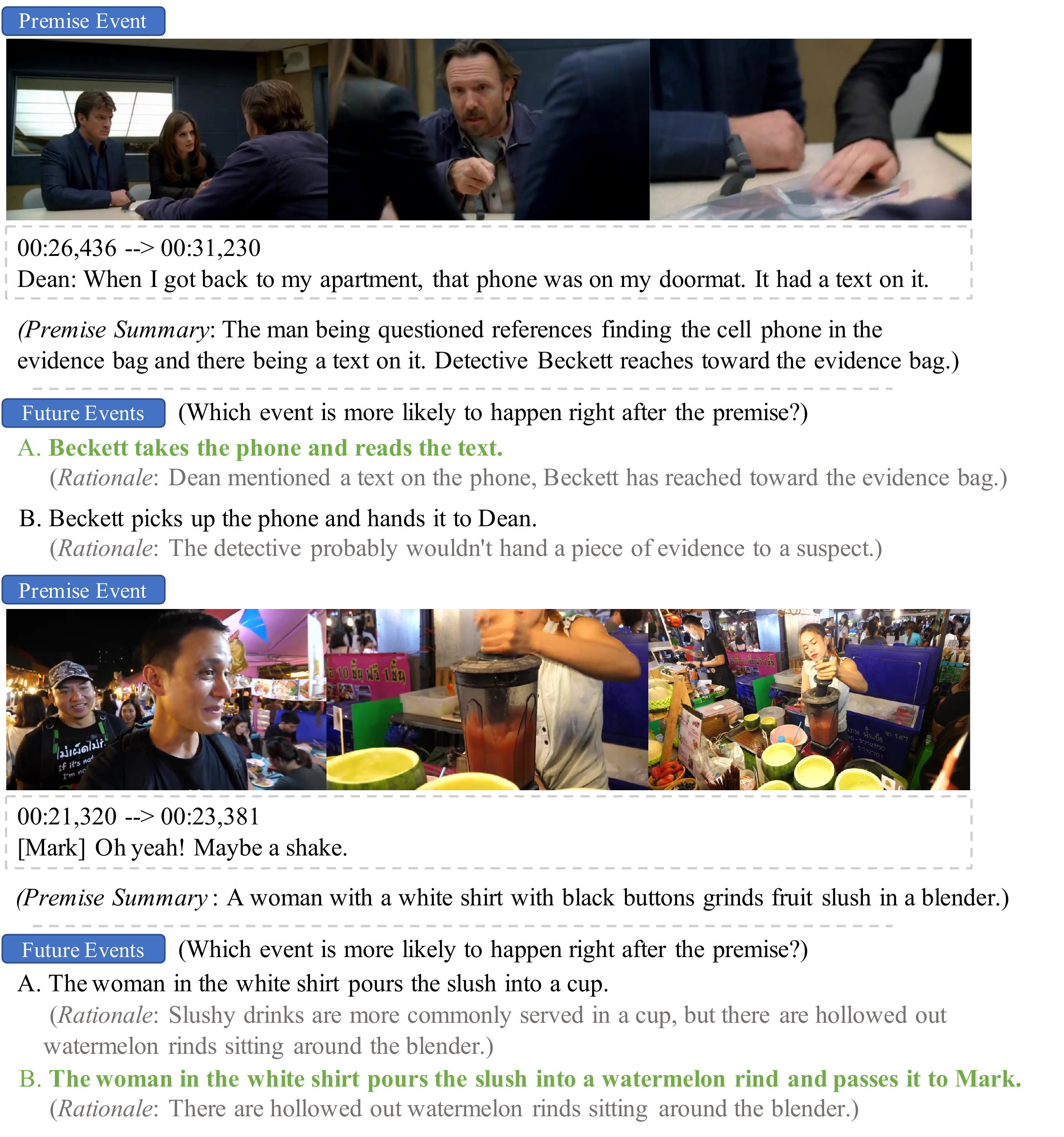}
  \caption{Video event prediction examples. Given a video (with dialogue) and two future events, the task is to predict which event is more likely to happen following the video.
  \textit{Top}: an example with a TV show clip. \textit{Bottom}: an example with a YouTube Lifestyle Vlog clip. The correct answer is shown in bold and green. \textit{Premise Summary} and \textit{Rationale} are included for illustration purpose only, they are hidden for the task.}
  \label{fig:data_example}
\end{figure}

Given a video clip (\textit{premise event}), humans can often describe logical events that might happen next (\textit{future events}), and interestingly people tend to agree on which future events are more likely to happen than others.
Making such predictions requires not only a deep understanding of the rich dynamics underlying the video and dialogue, but also a significant amount of \textit{multimodal commonsense} knowledge about the world.  
In Figure~\ref{fig:data_example} (\textit{top}), we show an example where commonsense knowledge about inter-human relationships is required, i.e., that a detective typically does not hand evidence to a suspect in a criminal case. 
Given this knowledge,
it is more likely that Beckett (the detective) will take the phone (the evidence) and read the text, than hand the phone to Dean (the suspect).

In this work, we propose \textit{Video-and-Language Event Prediction} (\dsetname), a novel dataset and task for fine-grained future event prediction from videos.
Given a video with aligned dialogue, and two possible future events, 
the AI system is required to understand both visual and language semantics from this video, and commonsense world knowledge, and then make a sound and practical judgment about the future, by choosing the more likely event from two provided possible future events.
The \dsetname~dataset contains 28,726 examples from 10,234 short video clips. 
Each example (see Figure~\ref{fig:data_example}) consists of a \textit{Premise Event} (a short video clip with dialogue), a \textit{Premise Summary} (a text summary of the premise event), and two potential natural language \textit{Future Events} (along with \textit{Rationales}) written by people. 
These clips are on average 6.1 seconds long and are harvested from diverse event-rich sources, i.e., TV show and YouTube Lifestyle Vlog videos.

Collecting such a dataset is a non-trivial task, as crowd-workers may write trivial negatives (less-likely events) that contain \textit{biases} or \textit{annotation artifacts}~\cite{gururangan2018annotation}, such as negation (e.g., \textit{`says nothing'}) or impolite actions (e.g., \textit{`hit someone in the face'}), as shown in Table~\ref{tab:annotation_artifacts}.
To mitigate this, we combine two recent effective approaches, adversarial human-and-model-in-the-loop data collection~\cite{nie2019adversarial} and adversarial matching~\cite{zellers2019recognition}, to build a larger, more-challenging, and less-biased dataset. 
Specifically, 50\% of the examples in \dsetname~are directly annotated by humans over two rounds: round one of standard data collection, i.e., crowd-workers perform the annotations with no model feedback, and round two of adversarial data collection, i.e., crowd-workers perform the annotations with the goal of fooling our basic models trained on round one data (thus avoiding obvious biases). 
Our analysis shows that the adversarial data collection helps to mitigate dataset bias (reduce trivial negatives), i.e., we notice that a premise-oblivious model (that does not see the premise event) performs worse on data collected on round two than that of round one.
Another 50\% of the examples are obtained by performing adversarial matching on the human-annotated positive events (more-likely events), i.e., for each premise event, we sample a positive from other premises as a negative, such that the sampled negative is relevant to the current premise while not being overly similar to the true positive. Overall, our dataset is collected via 3 methods (standard-human, adversarial-human, adversarial-matching), hence maintaining a balance between easy and hard examples while reducing potential biases.

To provide a strong baseline for this challenging multimodal future-prediction task, we propose a transformer-based model to incorporate both visual and textual information from the premise event.
We also inject commonsense reasoning knowledge into our model from the ATOMIC dataset~\cite{sap2019atomic}. 
Our ablation study shows that each part of our model, i.e., video understanding, dialogue understanding, and commonsense knowledge, is useful for the multimodal event prediction.
Though our model has shown promising results, it is still not comparable to human performance (67.46\% \textit{vs.} 90.50\%), indicating the challenging nature of the multimodal event prediction task and the large scope for interesting future work on our \dsetname~dataset.

To summarize, our contributions are 3-fold: (1) We propose a new task, Video-and-Language Event Prediction, which requires a model to make fine-grained, multimodal prediction regarding which future event is more likely to happen following a premise video. (2) We introduce a new dataset \dsetname~for the task, and use two approaches to gather natural hard-negative future-events: adversarial data collection and adversarial matching. This helps mitigate potential annotation artifacts and biases in the dataset. A detailed analysis of \dsetname~is provided. (3) We present a strong baseline method to benchmark the proposed dataset, and show that incorporating commonsense knowledge improves performance, indicating future directions for this new task (with a large model-human performance gap).

%% file: 2.related_work.tex
\section{Related Work}\label{sec:related_work}

\paragraph{Video-and-Language Understanding.} 
Various datasets and tasks have been introduced in this area, such as video captioning~\cite{xu2016msr,rohrbach2017movie,Wang_2019_ICCV,lei2020tvr}, video QA~\cite{tapaswi2016movieqa,jang2017tgif,Lei2018TVQALC}, and moment retrieval~\cite{anne2017localizing,gao2017tall,lei2020tvr}.
Recently, \citet{liu2020violin} propose the video-and-language inference task where a model needs to infer whether a statement is entailed or contradicted by a video.
While this task requires judging a statement's verification w.r.t. existing events, our task requires predicting future events.

\paragraph{Commonsense Reasoning.}
Recently, commonsense reasoning has emerged as an important topic in both language~\cite{zellers2018swag,zellers2019hellaswag,sap2019atomic} and vision~\cite{vedantam2015learning,zellers2019recognition,zadeh2019social,fang2020video2commonsense} communities. 
\citet{zellers2018swag,zellers2019hellaswag} build multiple-choice QA datasets for commonsense inference with text context, 
\citet{zellers2019recognition,park2020visual} propose datasets for commonsense-based QA and captioning on still images, \citet{fang2020video2commonsense} augment MSRVTT~\cite{xu2016msr} videos with commonsense captions and QAs.
In this work, we focus on a more complex type of context (video with dialogue) and a future prediction task, posing challenges for both video-and-dialogue understanding and commonsense reasoning .

\paragraph{Video Forecasting.}  Predicting the future is one of the popular research areas in the vision community. It covers a wide spectrum of topics, including predicting future frames~\cite{vondrick2016generating, liang2017dual}, future action labels~\cite{vondrick2016anticipating, gao2017red, Shi_2018_ECCV, epstein2020oops}, future human motions~\cite{Fragkiadaki_2015_ICCV, mao2019learning}, etc. While these works mostly study low-level vision or semantic concepts prediction (e.g., action labels), we focus on predicting high-level future events from video and dialogue.

\paragraph{Bias in Datasets.}
It is known that \textit{biases} or \textit{annotation artifacts}~\cite{goyal2017making,gururangan2018annotation,mccoy2019right,tsuchiya2018performance,poliak-etal-2018-hypothesis,zellers2019recognition} exist in standard human annotated datasets~\cite{bowman2015large,williams2017broad,antol2015vqa,tapaswi2016movieqa,jang2017tgif,kim2017deepstory,Lei2018TVQALC}.
For example, negation words such as \textit{nobody}, \textit{no} and \textit{never} are strong indicators of contradictions~\cite{gururangan2018annotation} in MNLI~\cite{williams2017broad}.
Such superficial patterns are easy for models to exploit, resulting in an overestimate of task performance~\cite{goyal2017making,gururangan2018annotation}.
\citet{zellers2019recognition} propose \textit{Adversarial Matching} to mitigate biases in QA, where positive answers are recycled to serve as negatives for other questions. 
\citet{nie2019adversarial} propose a \textit{Human-And-Model-in-the-Loop Entailment Training} (HAMLET) adversarial data collection strategy to gather challenging examples for NLI.
In this work, we adopt both approaches to construct a less-biased and more challenging dataset for the multimodal video+dialogue setting.

%% file: 3.dataset.tex
\section{Dataset}\label{sec:dataset}
The \dsetname~dataset contains 28,726 examples from 10,234 TV show and YouTube Lifestyle Vlog video clips. 
Of these, 50\% are collected directly from human annotators over two rounds: (1) round one: standard data collection; (2) round two: adversarial data collection. 
We collect human examples using Amazon Mechanical Turk (AMT), with an average cost of \$1.10 per example.
More detail about the annotators and quality checks are presented in Appendix Section~\ref{subsec:additional_data_collection_details}.
The other 50\% are obtained from human-annotated examples via Adversarial Matching~\cite{zellers2019recognition}. 
Hence, overall we build our dataset with 3 collection methods (standard-human, adversarial-human, adversarial-matching), allowing a balance between easy and hard examples while reducing potential biases.

\subsection{Video and Language Source}\label{subsec:video_collection}
\dsetname~is built using videos (with English dialogues) from two sources: TV shows and YouTube Vlogs. 
Both types of videos contain rich physical interactions and dialogues between people and are thus ideal sources for collecting interesting events.
We do not use videos from sources like ActivityNet~\cite{caba2015activitynet} since they do not have dialogues and typically contain fewer events.

\paragraph{TV Show Videos.}
We use TV show clips from TVQA~\cite{Lei2018TVQALC}. 
The clips are typically 60-90 seconds long, and are from 6 popular TV shows of 3 genres: 1) sitcom: \textit{The Big Bang Theory, How I Met Your Mother, Friends}, 2) medical drama: \textit{Grey's Anatomy, House}, 3) crime drama: \textit{Castle}.

\paragraph{YouTube Lifestyle Vlogs.}
While TV shows are good video sources with rich inter-human interactions, 
they may focus more on scripted content~\cite{lei2019tvqa+}.
Thus, we also collect a set of YouTube lifestyle vlogs as additional sources, which are typically more natural and live interactive.
We first manually identify a list of YouTube channels that contain videos with rich human interactions and dialogues (in English). 
We filtered out those channels with instructional videos~\cite{Miech_2019_ICCV} or routine videos~\cite{ignat2019identifying,fouhey2018lifestyle}, as they focus more on a single person performing actions, while we desire videos with richer multi-person interactions and dialogues. 
In addition, the actors in instructional or routine videos typically follow rigid steps (e.g., in cooking videos, they usually follow recipes) to finish a particular task, making it much easier to predict the future events.
In the end, we identified 9 channels that contain a diverse set of lifestyle vlog videos on various topics: \textit{travel}, \textit{food}, \textit{daily life} and \textit{family}, etc. 
We downloaded all videos from these channels that are published after 2017, which were then verified to ensure high quality.
The resulting pool contains 971 videos of 10-30 minutes long. 
Each video is associated with aligned dialogue text, either written by humans or generated from YouTube's Automatic Speech Recognition (ASR) system. 
We segment the videos into 60-second clips. 
For each video, we drop the first and the last clip, as most of them are high-level introductions or subscription pleas.

\subsection{Round One: Standard Data Collection}\label{subsec:round1}
As our task is video event prediction, we aim to collect a set of videos annotated with future event pairs (i.e., \textit{more-likely events} and \textit{less-likely events}, also referred to as \textit{positive events} and \textit{negative events}) that are likely to happen right after the `premise' video. 
Each event is written in natural language (English), and we require the positive event to be more likely to happen than the negative event.

With this goal in mind, we create our first annotation task on AMT.
We present workers (human \textit{writers}) with a 60-90 seconds long video with aligned dialogue subtitle, to encourage them to write events that are related to both the visual content and the dialogue.
Workers are required to first select an interesting event from the video with timestamps, similar to previous works~\cite{Lei2018TVQALC,lei2020tvr}.
This event is defined as the \textit{premise event}.
We also require workers to write a \textit{premise summary} -- a natural language (English) description summarizing the premise event.
Following~\citet{Lei2018TVQALC,lei2020tvr}, for referring a specific person in the video, workers are instructed to either use the character names (e.g., `Sheldon') if they are available in the dialogues or provide a referring expression~\cite{kazemzadeh2014referitgame} (e.g., `the man in blue top') that uniquely refers to a person in the video. 
Next, given the premise event, workers are required to write two \textit{future events}, one more likely  ($>$50\% chance) to happen after the premise event, and one less likely ($<$50\% chance).
For example, in Figure~\ref{fig:data_example}, the correct answers are the more-likely while the wrong answers are the less-likely.
To encourage workers to write more reasonable future event that ground to the premise event,\footnote{Otherwise, workers sometimes write random events that are not related to the given premise.} we also require them to provide a \textit{rationale} as to why it is more or less likely. 
As it is not the focus of this work, we will release these rationales to support research on textual explanation generation/classification tasks~\cite{huk2018multimodal,zellers2019recognition}.

Each collected example is verified by three human \textit{verifiers}, by ranking the future events conditioned on the premise event.
We only accept an example if at least three out of four users (one writer + three verifiers) reach an agreement, as~\citet{anne2017localizing,nie2019adversarial}.
In addition, we also discard examples if one of the verifiers thinks the events are against our instructions (e.g., wrong person reference).
In total, we collected 6,458 verified examples from 2329 TV show clips. 
We split them into 70\% training, 15\% development, and 15\% testing splits such that the videos and their corresponding examples only appear in one split.

\begin{figure*}[t]
  \centering
  \includegraphics[width=0.96\textwidth]{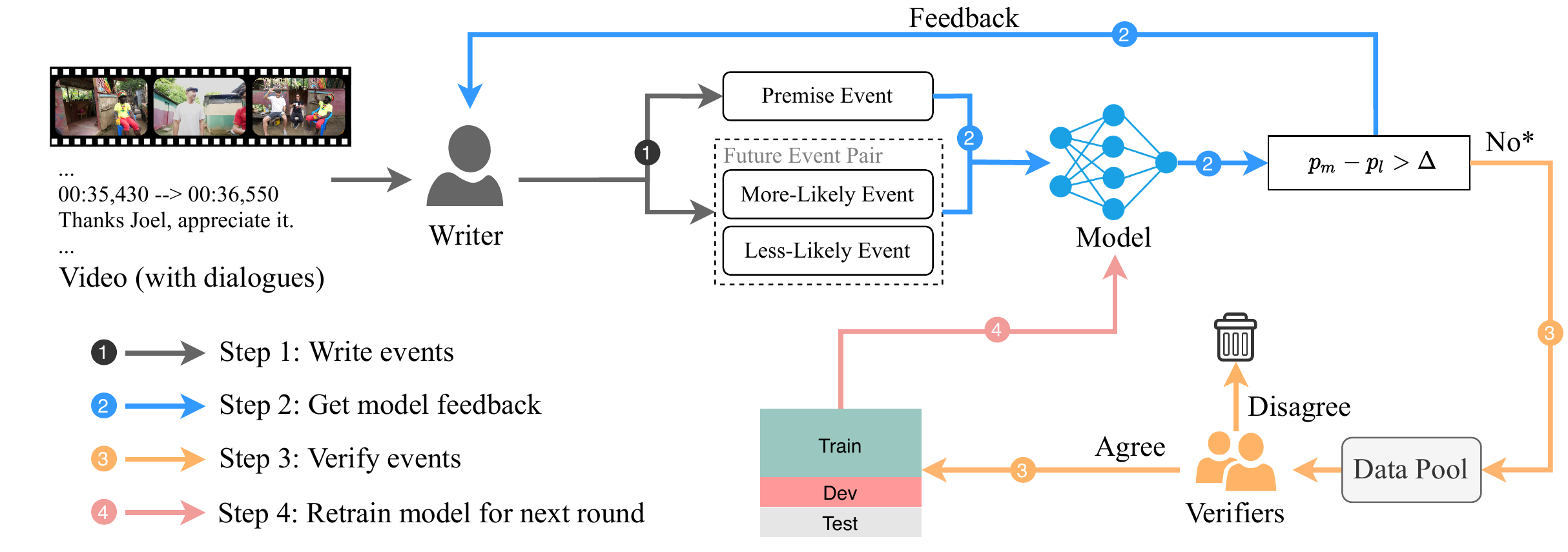}
  \caption{Illustration of our adversarial data collection procedure. $p_m$ and $p_l$ are the probabilities of the more-likely and the less-likely event being happening, respectively. $\Delta$ is a hyperparameter that controls how hard we want the collected example to be, it also helps to reduce prediction noises from imperfect models. $\Delta$ is set to 0.1 in our experiment. No* or the number of trials reaches the maximum limit of three. 
  }
  \label{fig:adversarial_collection_diagram}
\end{figure*}

\begin{table}[t]
\centering
\small
\scalebox{0.79}{
\begin{tabular}{l}
\toprule
\textbf{Type}: \textbf{Negation}\\
\textbf{Premise Summary}: Amy picks up her phone and reads a text message. \\
\textbf{More-likely}: Amy tells her friends what the text message says.\\
\textbf{Less-likely}: Amy says \textbf{nothing} at all to her friends.\\
\midrule
\textbf{Type}: \textbf{Impolite Actions}\\
\textbf{Premise Summary}: Chandler finds out that Joey used his toothbrush. \\
\textbf{More-likely}: Chandler starts arguing with Joey for using his toothbrush.\\
\textbf{Less-likely}: Chandler \textbf{hits Joey in the face with a punch.} \\
\bottomrule
\end{tabular}
}
\caption{Example annotation artifacts in the negative future events (\textit{Less-likely} events).}
\label{tab:annotation_artifacts}
\end{table}

\subsection{Round Two: Adversarial Data Collection}\label{subsec:round2}
While being efficient in data collection, we found the collected negative events in round one are sometimes simple and contain \textit{biases} or \textit{annotation artifacts}~\cite{gururangan2018annotation}.
In Table~\ref{tab:annotation_artifacts}, we show typical examples of annotation artifacts.
For example, we found workers tend to use negation when writing the less-likely event.
This particular type is similar to the \textit{visual priming bias}~\cite{zhang2016yin} for \textit{yes/no} questions in VQA~\cite{antol2015vqa} and the \textit{negation word bias}~\cite{gururangan2018annotation} in MNLI~\cite{williams2017broad}. 
To quantitatively study the effect of these annotation artifacts, we fine-tune a RoBERTa-base~\cite{liu2019roberta} model to classify which event is more likely to happen, with only the future events from round one's training data, i.e., the model has no access to the premise event.
On round one's Dev. split, this premise-oblivious model obtains 75.34\% accuracy, which is much higher than chance (50\%).

Hence, in order to collect harder and less-biased negatives, we make use of an adversarial collection procedure (see Figure~\ref{fig:adversarial_collection_diagram}), in a human-and-model-in-the-loop process~\cite{nie2019adversarial}, where models are used to provide real-time feedback to crowd-workers during data collection. 
Specifically, each submitted result is sent to the model for evaluation and 
writers are prompted\footnote{Rewrite for at most twice, in total three trials.} to rewrite their negative event if our model predicts a much higher probability for the more-likely event ($p_m$) than the less-likely event ($p_l$), i.e., $p_m - p_l > \Delta$, where 
$\Delta$ is a hyperparameter that controls how difficult we want the collected examples to be and is set to $0.1$ empirically.
This can be seen as a \textit{soft-adversarial} strategy, unlike~\citet{nie2019adversarial} where feedback decisions are made by directly using \textit{hard} model predictions (consider it as a special case of our soft-adversarial strategy with $\Delta = 0$). 
In addition to controlling the difficulty of the collected examples, it also helps us to reduce the prediction noise from imperfect models and avoid forcing workers to write abnormal events in order to fool the model.

We use two models to provide feedback to the writers, a \textit{future event only} model that focuses primarily on reducing the aforementioned annotation artifacts, and a \textit{premise summary + future event} model that can additionally detect and thus reduce simple negatives that are created as contradictions of the premise.
For example, with the premise summary, \textit{`Howard tells Bernadette that he has a dominant personality'}, the negative event \textit{`Howard will say that he doesn't have a dominant personality'} is relatively simple as it directly contradicts the premise.
Both models are fine-tuned as a sequence classification task from round one's training data, using a pre-trained RoBERTa-base\footnote{Empirically, RoBERTa-large does not yield better performance but longer response time that affects user experience.} model.
The objective is to maximize the probability of the positive event being the correct answer.
For the future event only model, we only use the future event for classification, ignoring the premise. 
For the premise summary + future event model, we concatenate the premise summary and the future event text as a single sequence for classification. 
Note that we use the premise summary as an overall proxy to represent both video and dialogue content to build our adversarial model, considering video and dialogue understanding is still an open research problem in itself.\footnote{In Appendix Section~\ref{subsec:more_results}, we show that an oracle model that uses the premise summary as auxiliary input significantly outperforms our video+dialogue model.}
The accuracy of these two models on round one's Dev. split are 75.34\% and 76.68\%, respectively.
During collection, we randomly pick one model from these two models to provide feedback to users.
This is similar to the approach used by~\citet{nie2019adversarial} where one model is randomly picked from a set of random seeded models. 
The difference lies in that we use a set of two models with different inputs (architecture) while~\citet{nie2019adversarial} use the same architecture with varying random seeds.
This strategy can be seen as constructing a pseudo-ensemble model, which provides diverse adversarial feedback to the crowd-workers and helps avoid annotators exploiting vulnerabilities of a single model~\cite{nie2019adversarial}, while reducing server load.\footnote{As we only need to run one model instead of multiple models in a standard ensemble approach.}

In round two, with our adversarial collection procedure, we collected 7,905 verified examples from 4,418 TV show clips and 3,487 YouTube clips. Similar to round one, we split them into 70\% training, 15\% development, and 15\% testing splits.

\subsection{Adversarial Matching}\label{subsec:adv_matching}
With adversarial data collection, we are able to collect harder and less-biased examples. 
However, this approach is not scalable due to its high cost. 
On average, each verified example in round two costs \$1.70.
Inspired by~\citet{zellers2019recognition}, which proposed to use Adversarial Matching to obtain less-biased negatives, we use a similar strategy to create additional examples for our dataset.
Given a premise event and its positive event, the goal of adversarial matching is to find a negative from other premise events' positives, such that the matched negative is very relevant to the premise event (so that they are still hard for machines) and at the same time, not overly similar to the true positive (in case they incidentally become a positive event to the premise). 
Specifically, we use BERTScore~\cite{zhang2019bertscore} and the recommended RoBERTa-Large model fine-tuned on MNLI~\cite{williams2017broad} to calculate similarity score $S_{sim}(e_i, e_j)$ between two events $e_i$ and $e_j$. 
For relevance, we use a RoBERTa-base model that takes as input the concatenation of a premise summary $p_i$ and a future event $e_j$  and output a relevance score $S_{rel}(p_i, e_j)$.
This model is trained to distinguish positive events from randomly sampled events.
Next, given dataset examples $\{(p_i, e_i)\}_{i=1}^{N}$, we obtain a negative future event for each premise $p_i$ with maximum-weight bipartite matching~\cite{munkres1957algorithms,jonker1987shortest} on a weight matrix $\mathbf{W} \in \mathbb{R}^{N \times N}$:
\begin{align*}
\mathbf{W}_{i,j} &=\lambda (S_{rel}(p_i, e_j) - \alpha S_{sim}(e_i, e_j)), \\
\lambda &= (1 - 0.5 \cdot \mathbb{1}(p_i, e_j)), 
\end{align*}

\noindent
where $\alpha \mbox{=} 0.1$ is a hyperparameter that controls the tradeoff between relevance and similarity, the indicator $\mathbb{1}(p_i, e_j)$ equals 1 if $p_i$ and $e_j$ are from different sources (e.g., different TV shows), otherwise 0. 
Thus, $\lambda$ serves as a regularization that penalizes $e_j$ if it is from a different video source than that of $p_i$ -- as $e_j$ could potentially be an easy negative that can be distinguished from superficial clues such as character names in different shows.

\begin{table}[t]
\centering
\small
\setlength{\tabcolsep}{4pt}
\scalebox{0.78}{
\begin{tabular}{lrcccr}
\toprule
\multirow{2}{*}{ Split } & \multirow{2}{*}{ \#Videos } & Pre. Event & \multicolumn{2}{c}{ Avg. Sen. Len. (\#words) } & \multirow{2}{*}{ \#Examples } \\ \cmidrule(){4-5}
& & Avg. Len. (s) & Pre. Sum. & Pos. / Neg. & \\
\midrule
Train & 7,180 & 6.1 & 15.2 & 11.1 / 11.2 & 20,142 \\
Dev & 1,561 & 6.2 & 14.7 & 11.0 / 11.1 & 4,392 \\
Test & 1,493 & 6.2 & 15.4 & 11.0 / 11.1 & 4,192 \\
\midrule
Total & 10,234 & 6.1 & 15.2 & 11.1 / 11.2 & 28,726 \\
\bottomrule
\end{tabular}
}
\caption{Statistics by Data Split. \textit{Pre. Event}=Premise Event, a short video with dialogue. \textit{Pre. Sum.}=Premise Summary. \textit{Pos. }/\textit{Neg.}=Positive/Negative future event.}
\label{tab:data_stat_split}
\end{table}

\begin{table}[t]
\centering
\small
\setlength{\tabcolsep}{4pt}
\scalebox{0.82}{
\begin{tabular}{llrrr}
\toprule
\multirow{2}{*}{Domain} & \multirow{2}{*}{Genre} & \#Shows \ \ \ & \multirow{2}{*}{\#Videos} & \multirow{2}{*}{\#Examples} \\
 &  & (\#Channels) &  &  \\
\midrule
\multirow{3}{*}{ TV show } & Sitcom & 3 & 4,117 & 12,248 \\
& Medical & 2 & 1,558 & 5,198 \\
& Crime & 1 & 1,072 & 4,306 \\
\midrule
\multirow{2}{*}{ YouTube Vlogs } & Travel, Food & 6 & 2,406 & 4,812 \\
& Family, Daily & 3 & 1,081 & 2,162 \\
\midrule
Total & - & 15 & 10,234 & 28,726 \\
\bottomrule
\end{tabular}
}
\caption{Data Statistics by Genre.}
\label{tab:data_stat_by_genre}
\end{table}

\begin{table}[t]
\centering
\small
\setlength{\tabcolsep}{4pt}
\scalebox{0.96}{
\begin{tabular}{ll}
\toprule
Genre & Top Unique Verbs \\
\midrule
\multirow{2}{*}{Sitcom} & change, offer, hear, should, accept, yell,   \\
 &  hang, join, apologize, shut, shout, realize\\
\midrule
\multirow{2}{*}{Medical} & die, treat, cry, yell, smile, proceed, examine, \\
 & approach, argue, save, admit, rush\\
\midrule
\multirow{2}{*}{Crime} &  kill, shoot, point, question, toss, hang, \\
 &  remove, catch, lie, deny, investigate,\\
\midrule
\multirow{2}{*}{Travel, Food} &  taste, add, pour, dip, cook, describe, cut, \\
 &  order, serve, stir, prepare, enjoy, buy\\
\midrule 
\multirow{2}{*}{Family, Daily} &  drive, jump, wear, point, smile, touch,  \\
 &   climb, dress, set, swim, hide, lay, blow\\
\bottomrule
\end{tabular}
}
\caption{Top unique verbs in each genre.}
\label{tab:top_unique_verb}
\end{table}

\subsection{Data Analysis}
Table~\ref{tab:data_stat_split} shows the overall statistics of the dataset and data splits details.
Each example in our dataset is paired with a premise event clip, with an average length of 6.1 seconds.
The average length of our positive event (\textit{Pos.}) sentences is very close to that of the negative (\textit{Neg.}) ones (11.1 \textit{vs.} 11.2), suggesting little bias in sentence length. 
Our videos are curated from TV shows and YouTube vlogs, across five major categories with diverse topics, i.e., \textit{sitcom, medical, crime, travel-food, family-daily}. 
In Table~\ref{tab:data_stat_by_genre} we show data statistics by genre. 
Events generally vary by genre. To demonstrate these differences, we show top unique verbs in each genre in Table~\ref{tab:top_unique_verb}. 
The top unique verbs in \textit{Crime} genre are usually close to crime and violence, while top unique verbs in \textit{Family, Daily} are usually related to daily activities such as `drive' and `wear'.
For top unique nouns and additional data analysis (e.g., distribution of examples by reasoning type), please see Appendix Section~\ref{subsec:additional_data_analysis}.
For adversarial data collection in round two, the average number of trials is 2.7, 
i.e., on average the writer has to write their negative event for 2.7 times.
For the first trial, 59.21\% of the examples are defined as \textit{easy} by our system, i.e., the positive event has a much larger probability of happening than the negative event. 
With rewriting, only 31.22\% of the examples remain \textit{easy}.
Moreover, in Table~\ref{tab:acc_by_data_source} row 1, when trained on our final dataset, we show that our future event only baseline gets much lower performance on the round two subset than that of round one (59.62\% \textit{vs.} 74.20\%), showing round two examples are less-biased.

%% file: 4.method.tex
\section{Method}\label{sec:method}
\begin{figure}[!t]
    \centering
  \includegraphics[width=0.98\linewidth]{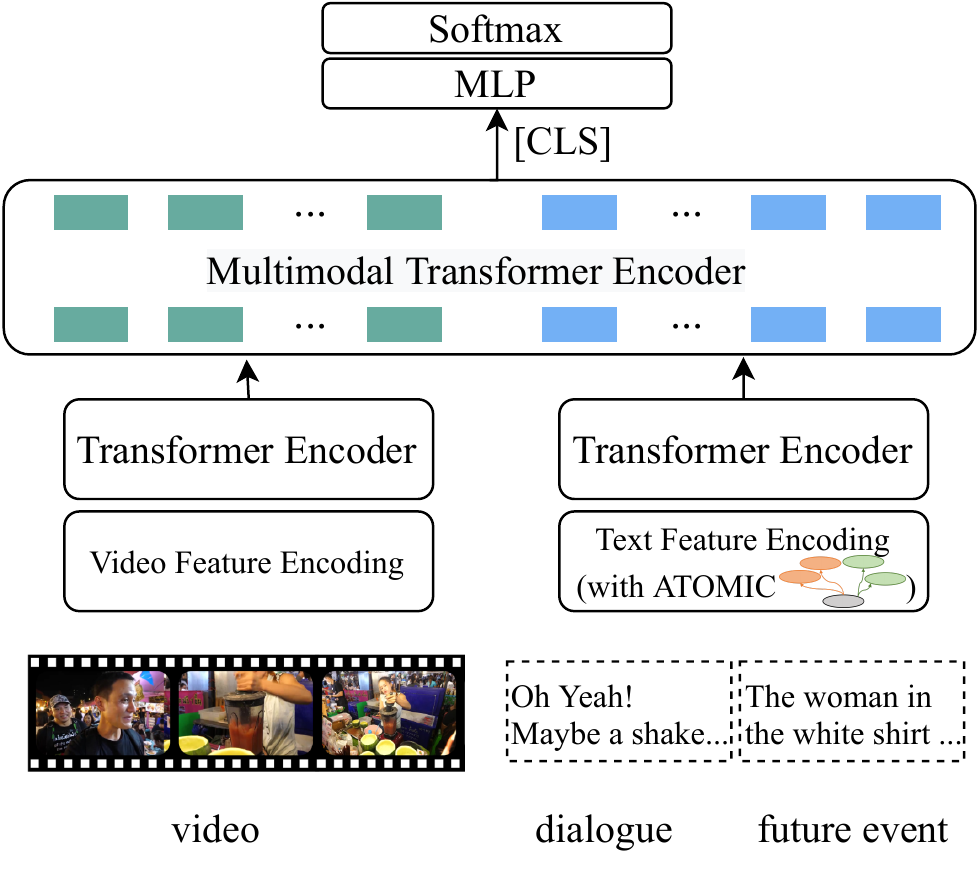}
  \vspace{-5pt}
  \caption{Model overview. We first separately encode video and text, and then use a multimodal transformer encoder to encode information from both modalities. Please see text for details.}
  \label{fig:model}
\end{figure}

Given a video with dialogue text, and two future event candidates $\{e_i\}, i \in \{1, 2\}$, our goal is to predict which future event is more likely to happen. 
In the following, we introduce our baseline approach (see model overview in Figure~\ref{fig:model}) for this new task.

\paragraph{Video Encoding.} 
We encode each video using appearance and motion features at 1 FPS. For appearance, we extract 2048D feature vectors from the ImageNet~\cite{deng2009imagenet} pre-trained ResNet-152~\cite{he2016deep}. 
For motion, we extract 2048D feature vectors from the Kinetics~\cite{carreira2017quo} pre-trained ResNeXt-101~\cite{hara2018can}. 
These features have shown to perform well in several video and language tasks~\cite{Miech_2019_ICCV}.
We perform L2-normalization and concatenate the features as the video representation. We project these representations into a lower dimension space and add a trainable positional encoding~\cite{devlin2018bert} to them. We then use a \textit{transformer encoder}~\cite{vaswani2017attention} to further encode the resulting representation, denoted as $E^{v} \in \mathbb{R}^{T \times d}$.

\begin{figure}[!t]
    \centering
  \includegraphics[width=0.96\linewidth]{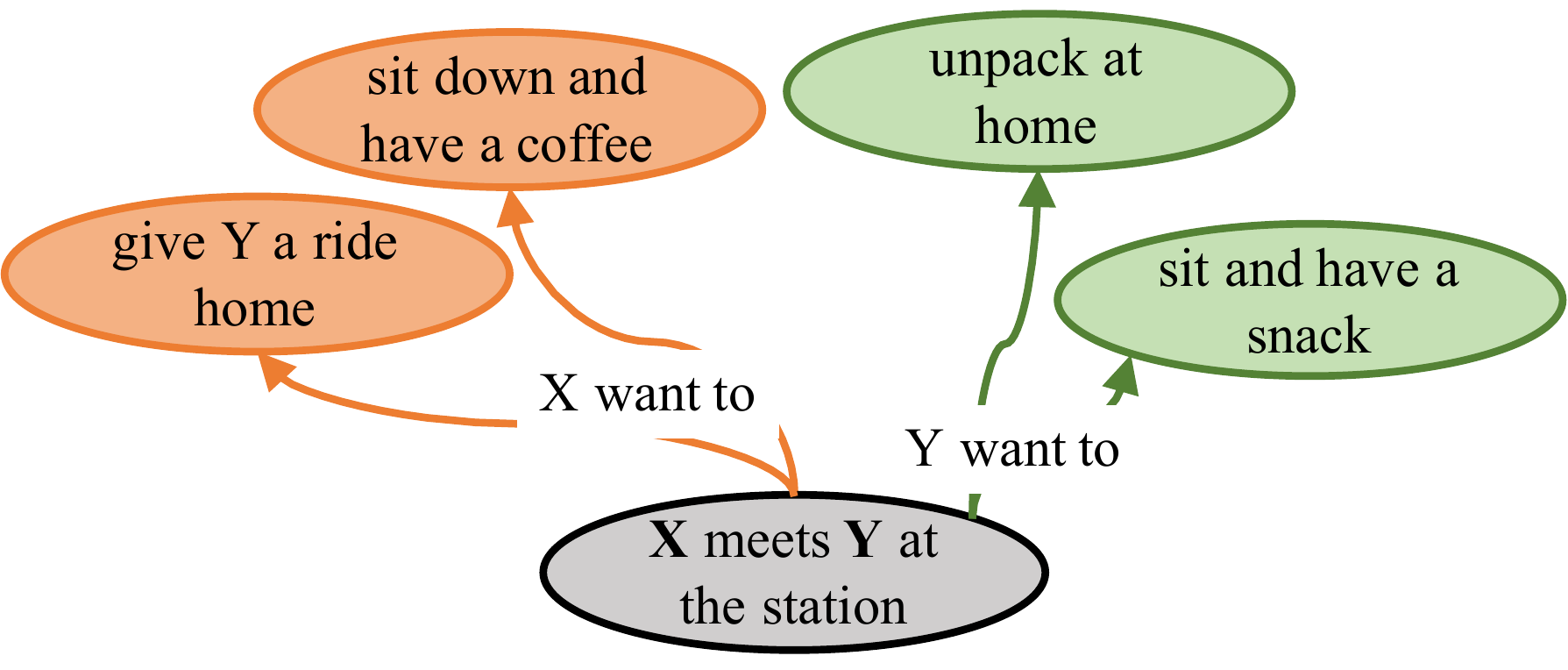}
  \caption{An ATOMIC~\cite{sap2019atomic} example.}
  \label{fig:atomic_graph}
\end{figure} 

\paragraph{Text Encoding.}  
For text, we use the contextualized text features from the RoBERTa-base~\cite{liu2019roberta}. 
We first fine-tune the pre-trained RoBERTa with commonsense knowledge extracted from the ATOMIC dataset~\cite{sap2019atomic} (see details in the paragraph below) and then use the resulting model for feature encoding. Note that this model is end-to-end trainable during training.
We concatenate dialogue and future event candidate as input to the transformer layers, special tokens such as {\tt [CLS]}~\cite{devlin2018bert} are also added in this process. 
We use the extracted token embeddings from the last layer, denoted as $E_{i}^{t} \in \mathbb{R}^{L_i \times d}, i \in \{1, 2\}$, where $L_i$ is the sequence length (\#tokens, including added special tokens). 
Similar to how we encode video, the resulting text representation is further encoded using a transformer encoder. 
Without ambiguity, we use the same notation to denote the outputs as $E_{i}^{t} \in \mathbb{R}^{L_i \times d}, i \in \{1, 2\}$.

\paragraph{Commonsense-based Text Representations.} \
Addressing our challenging future event prediction task requires general world knowledge that is beyond basic visual and language semantic understanding. 
Thus, we propose to inject the commonsense from the ATOMIC dataset~\cite{sap2019atomic} into our framework in a simple way.
ATOMIC contains events with if-then inferences, e.g., \textit{if X meets Y at the station, then X want to give Y a ride home} (see example in Figure~\ref{fig:atomic_graph}).
We extract 406K event inferences from the dataset, and replace the person tokens \textit{X} and \textit{Y} with the names from our dataset~\cite{Mitra2019ExploringWT}.
We then use the extracted event inference sentences to finetune the pre-trained RoBERTa-base model.
The fine-tuned model is then used to encode our text inputs.

\paragraph{Multimodal Encoding and Event Classification.}
To obtain the joint multimodal representation, we concatenate encoded video $E^v$ and text $E^t$ and use a transformer encoder to encode the concatenated representations. 
This encoder allows information exchange between the two modalities. 
We use the representation from the {\tt [CLS]} token as the joint representation of video, dialogue and future event $e_i$, denoted as $g_i \in \mathbb{R}^{d}, i \in \{1, 2\}$.
We gather the joint representation vectors for all future event candidates and pass them through a two-layer MLP with a softmax layer for classification.
We train the model using cross-entropy loss that maximizes the scores for the more-likely future events.

%% file: 5.experiments.tex
\section{Experiments}\label{sec:exp}

\subsection{Implementation Details} 
Our models are implemented in PyTorch~\cite{paszke2017automatic}. To speed up training, we use NVIDIA Apex for mixed precision training. 
We set the hidden size $d$ to be 768 and use a single transformer layer for all our transformer encoders.
We use Adam~\cite{kingma2014adam} optimizer with $\beta_1\mbox{=}0.9$, $\beta_2\mbox{=}0.999$. 
Since our model has a pre-trained component (RoBERTa), we use a two-phase training strategy. Specifically, we first freeze RoBERTa's weights up to the second last layer and then pre-train the rest of model for 3 epochs with initial learning rate of 1e-4, learning rate warmup over the first 10\% of the steps and linear decay the learning rate to 0. We then unfreeze all the weights and finetune the whole model for 3 epochs with learning rate 5e-5 and linearly decay the learning rate to 0.
We train the model on a single RTX 2080Ti GPU with batch size 16.
We report multiple-choice question answering accuracy.

\subsection{Results}

\begin{table}[t]
\centering
\small
\setlength{\tabcolsep}{4pt}
\begin{tabular*}{\linewidth}{l@{\extracolsep{\fill}}r}
\toprule
Model & Accuracy (\%)\\
\midrule
chance & 50.00  \\
future only  & 58.09 \\
video + future &  59.03\\
dialogue + future & 66.63 \\
video + dialogue + future  & 67.46 \\
\midrule
human (dialogue + future) & 76.25 \\
human (video + dialogue + future)& 90.50 \\
\bottomrule
\end{tabular*}
\caption{Results on \dsetname~Test split.}
\label{tab:main_res_table}
\end{table}

\begin{figure*}[t]
  \centering
  \includegraphics[width=0.95\textwidth]{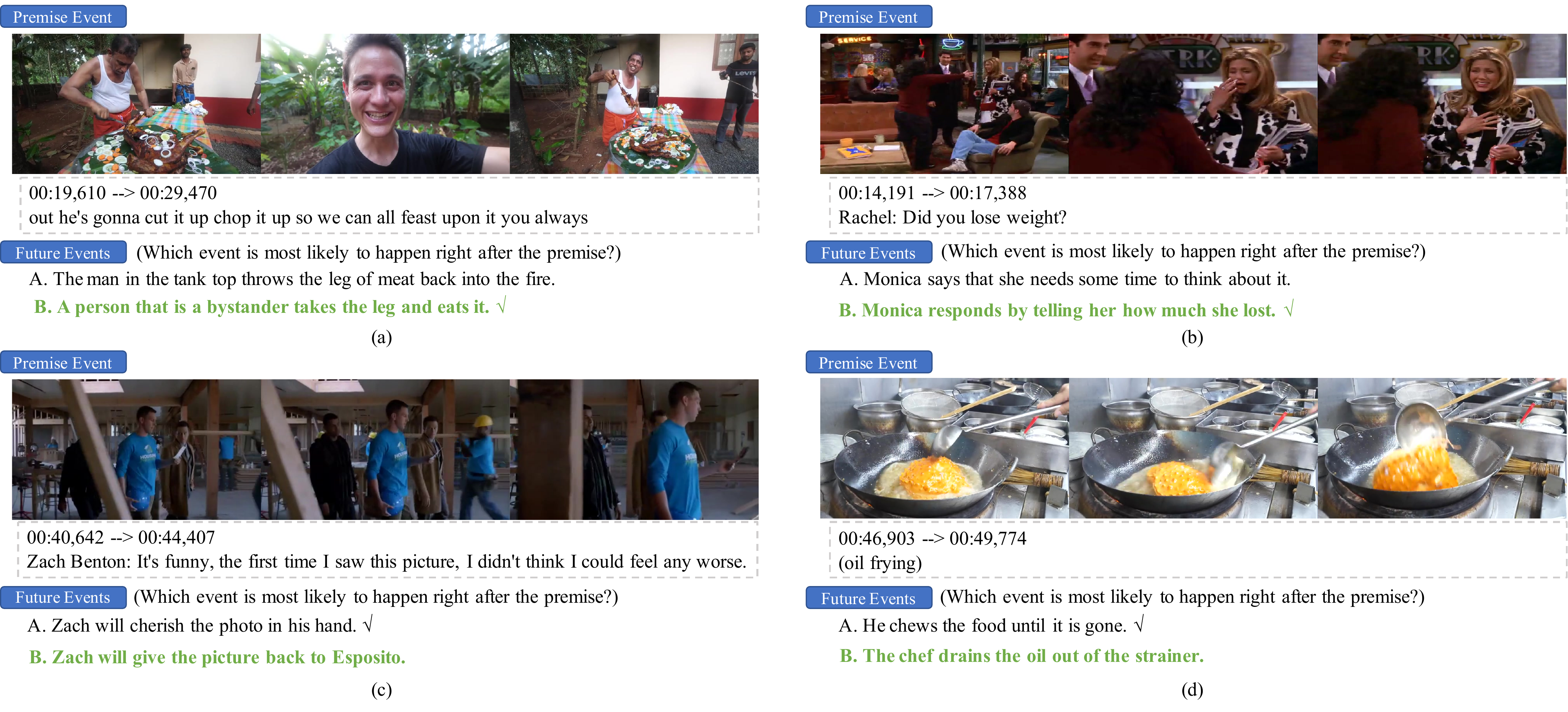}
  \caption{Prediction examples from our best model. Top row shows correct predictions, bottom row shows failure cases. Left column shows human annotated examples, right column shows adversarial matching examples. Ground truth answers are in bold and green, model predictions are indicated by \checkmark.}
  \label{fig:model_pred_examples}
\end{figure*}

\paragraph{Are video and dialogue modalities useful?} Table~\ref{tab:main_res_table} shows the results with different input context. 
The model using future event text only as the input achieves 58.09\% accuracy, which is higher than random chance (50\%), suggesting there exists slight bias even with our deliberate adversarial collection and matching but is tolerable.
Adding video or dialogue as additional input improves the accuracy to 59.03\% and 66.63\%, respectively.
The best performance is achieved when using both video and dialogue, with an accuracy of 67.46\%.
In Appendix Section~\ref{subsec:more_results}, we also present an oracle model with premise summary as auxiliary input.

\paragraph{Human Performance.} To obtain human performance, we randomly sampled 400 examples from our test set.
We present a premise event (a video with dialogue subtitles or dialogue subtitles only) and its two corresponding future events to a new set of AMT workers and ask them to select which one is more likely to happen after the premise. 
Each example is answered by 10 different workers to reduce crowdworker variance~\cite{rajpurkar2018know}. 
The final answer is selected by majority vote.
Table~\ref{tab:main_res_table} shows the results.
We observe that human performance without video (i.e., only dialogue+future) is 76.25\%, while showing the video improves the performance to 90.5\%. which indicates video information is important for getting the correct answer. 
Compared with the best model result (67.46\%), there is still a large useful gap (23\%) for future community work on our challenging task of multimodal event prediction.

\begin{table}[t]
\centering
\small
\setlength{\tabcolsep}{4pt}
\begin{tabular*}{\linewidth}{l@{\extracolsep{\fill}}r}
\toprule
Model & Accuracy (\%) \\
\midrule
video + dialogue + future  & 67.46 \\
\midrule
\quad - ATOMIC fine-tuning & 66.96 \\
\bottomrule
\end{tabular*}
\caption{Effect of ATOMIC fine-tuning.}
\label{tab:ablation}
\end{table}

\paragraph{Does commonsense knowledge help?} In Table~\ref{tab:ablation}, we show a model variant that uses text features without ATOMIC sentences for fine-tuning. 
We see this variant has a lower accuracy (66.96\%) compared with the fine-tuned accuracy (67.46\%).

\begin{table}[t]
\centering
\small
\setlength{\tabcolsep}{4pt}
\scalebox{0.76}{
\begin{tabular}{lrrrr}
\toprule
\multirow{2}{*}{Model} & Adv. Matching   & \multicolumn{2}{c}{Human-Annotated}  & Overall \\ \cmidrule{3-4}
 & (50\%)  & R1 (22\%) & R2 (28\%) & (100\%)  \\
\midrule
future only & 50.00 & 74.20 & 59.62 & 58.09 \\
video + future & 54.34 & 69.21 & 59.19 & 59.03 \\
dialogue + future & 67.60 & 70.70 & 61.53 & 66.63 \\
video + dialogue + future  & 68.37 & 70.59 & 63.26 & 67.46 \\
\bottomrule
\end{tabular}
}
\caption{Performance breakdown by data collection method.}
\label{tab:acc_by_data_source}
\end{table}

\begin{table}[t]
\centering
\small
\setlength{\tabcolsep}{5pt}
\begin{tabular*}{0.9\linewidth}{l@{\extracolsep{\fill}}rrrr}
\toprule
word & throws & face & leave & without \\
\midrule
PMI (R1) \quad \quad & 1.38 & 1.28 & 1.25 & 1.23 \\
PMI (R2) \quad \quad & 0.83 & 0.67 & 0.66 & 0.81 \\
\bottomrule
\end{tabular*}
\caption{Top words by PMI in standard collection (R1) and their values in adversarial collection (R2). The values are calculated using PMI(word, less-likely).}
\label{tab:pmi}
\vspace{-9pt}
\end{table}

\paragraph{Impact of Data Collection Method.} 
Table~\ref{tab:acc_by_data_source} shows the model performance breakdown by different collection methods.  For human-annotated data, we show performance on round one (\textit{R1}, standard data collection) and round two (\textit{R2}, adversarial data collection).
First, we observe that the accuracy of the \textit{future only} model matches chance on adversarial matching data while being higher on human-annotated data.
The main reason is the matched data has less artificial biases than human-annotated ones.
Second, for human-annotated data, across all models, we see the performance on round two subset is significantly lower than that of round one, which demonstrates the effectiveness of using our adversarial collection procedure. 

\citet{gururangan2018annotation} shows lexical choice is a strong indicator of the inference class in NLI. 
To check how our adversarial collection affects the use of words, we use pointwise mutual information (PMI) as in~\citet{gururangan2018annotation}.
In Table~\ref{tab:pmi} we show top words that are associated with negative class (less-likely event) in standard collection versus their values in our adversarial collection process. 
We find that the PMI values of these top negative words (e.g., `throws`, `without`, that frequently occur in negative less-likely events) in standard collection clearly drop in adversarial collection, e.g., `throws` drops from 1.38 to 0.83, making it less indicative of the negative.

\paragraph{Qualitative Examples.} 
We show 4 prediction examples using our best model (video + dialogue + future) in Figure~\ref{fig:model_pred_examples}.
Top row shows two correct prediction examples, where our model is able to predict basic human intention and reaction.
Bottom row shows two incorrect predictions, where wrong predictions are mainly caused by the lack of commonsense. 
For example, to correctly pick the more likely event in Figure~\ref{fig:model_pred_examples}(c), the model needs to understand that the `photo' is an evidence of a police investigation.
Figure~\ref{fig:model_pred_examples}(d) shows an example that requires the model to infer that the food is not ready yet. More examples are presented in Appendix Section~\ref{subsec:more_qualitative_examples}.

%% file: 6.conclusion.tex
\section{Conclusion}\label{sec:conclusion}

We introduce a new task, Video-and-Language Event Prediction (\dsetname) - given a video with aligned dialogue, and two future events, an AI system is required to predict which event is more likely to happen. 
To support this task, \dsetname~dataset is collected.
We present a strong transformer-based baseline that incorporates information from video, dialogue, and commonsense knowledge, each of which is necessary for this challenging task.

\section*{Acknowledgements} 
We thank the reviewers for their helpful feedback. This research is supported by NSF Award \#1562098, DARPA MCS Grant \#N66001-19-2-4031, DARPA KAIROS Grant \#FA8750-19-2-1004, ARO-YIP Award \#W911NF-18-1-0336, and Google Focused Research Award. The views contained in this article are those of the authors and not of the funding agency.

%% file: 7.appendix.tex
\section{Appendices}

\subsection{Additional Data Analysis}\label{subsec:additional_data_analysis}

Our videos are curated from two sources, TV shows and YouTube lifestyle vlogs, across five major categories, i.e., \textit{sitcom, medical, crime, travel-food, family-daily}. 
Events generally vary by genre. 
One way to show the difference is by checking the top unique nouns in each genre.
To obtain the top unique nouns, we first tokenize and lemmatize the future event sentences. 
Each resulting token is also tagged with a part-of-speech tag.
Next, for each genre, we take the top unique nouns as the ones among the most frequent 100 nouns from one genre but do not appear in those from the other genres combined. 
We show the top unique nouns in each genre in Table~\ref{tab:top_unique_noun}.
Interestingly, the top unique nouns in \textit{crime} genre are closer to crime and violence, while in \textit{family-daily}, the top unique nouns are relatively more family relevant.

\begin{table}[t]
\centering
\small
\setlength{\tabcolsep}{4pt}
\scalebox{0.96}{
\begin{tabular}{ll}
\toprule
Genre & Top Unique Nouns \\
\midrule
\multirow{2}{*}{Sitcom} & apartment, group, couch, bottle, game, date,\\
 &  joke, kitchen, story, wine, seat, hug\\
\midrule
\multirow{2}{*}{Medical} & patient, doctor, office, surgery, parent, \\
 &  elevator, hospital, nurse, team, case, cane\\
\midrule
\multirow{2}{*}{Crime} &  gun, picture, photo, paper, information, \\
 &   evidence, police, case, suspect, ground\\
\midrule
\multirow{2}{*}{Travel, Food} &  host, meat, bite, plate, bowl, chef, piece,\\
 &  sauce, fish, dish, soup, noodle, spoon\\
\midrule 
\multirow{2}{*}{Family, Daily} &  kid, dad, child, dog, son, toy, father,\\
 &   daughter, family, wife, video, candy, hair\\
\bottomrule
\end{tabular}
}
\caption{Top unique nouns in each genre.}
\label{tab:top_unique_noun}
\end{table}

\begin{figure}[t]
    \centering
  \includegraphics[width=\linewidth]{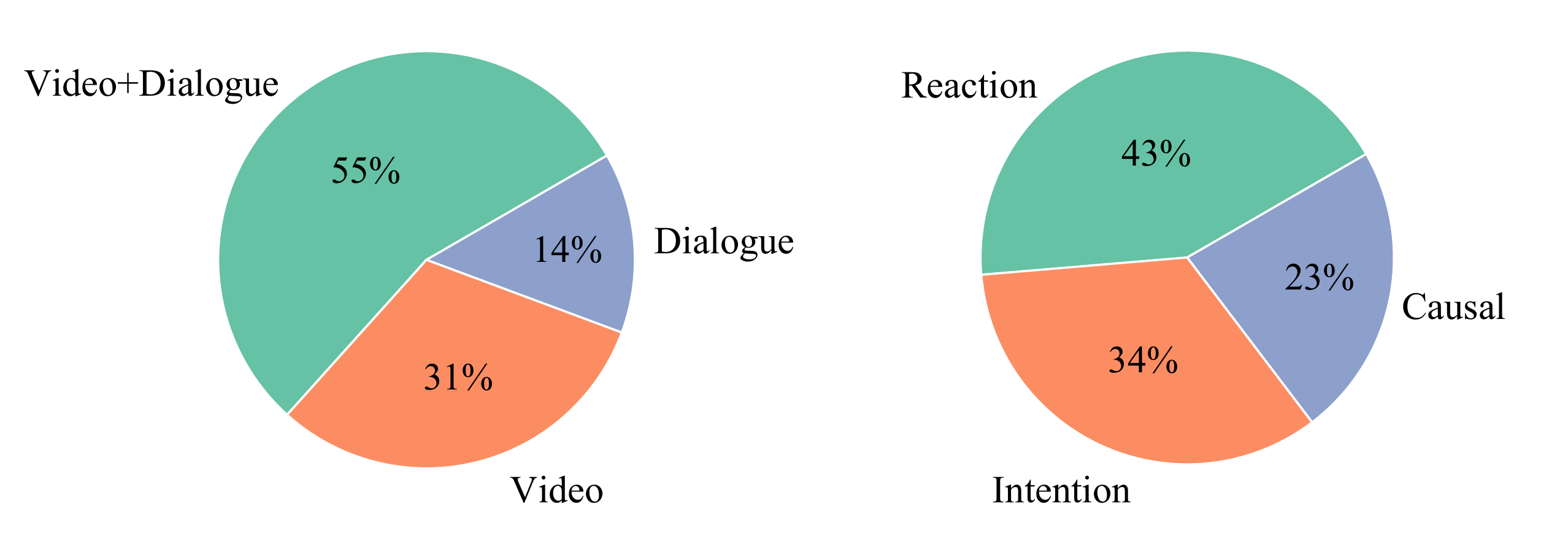}
  \vspace{-15pt}
  \caption{Distribution of examples by premise understanding type (\textit{left}) and by reasoning type (\textit{right}).}
  \label{fig:modality_dist_pie}
\end{figure}

Figure~\ref{fig:modality_dist_pie} (left) shows the distribution of examples by premise understanding type, i.e., what modalities are needed to understand the premise event. 
Most of the premise events require both video and dialogue understanding. 
Figure~\ref{fig:modality_dist_pie} (right) shows the distribution of examples by commonsense reasoning type.
We categorize commonsense reasoning into three types by examining the relation between the premise event and the positive future event: (1) intention, e.g., if \textit{X brings two cups of coffee}, then \textit{X (intends to) give Y a cup of coffee}. (2) reaction, e.g., if \textit{X hands Y a form and describes a procedure}, then \textit{Y signs the form and hands it back}. (3) causal, e.g., if \textit{X says they hit a bump}, then \textit{X gets unbalanced and falls off the boat}. 
The two distributions are obtained by manually annotating 100 randomly sampled examples from~\dsetname~Dev. split.

\begin{figure}[t]
    \centering
  \includegraphics[width=\linewidth]{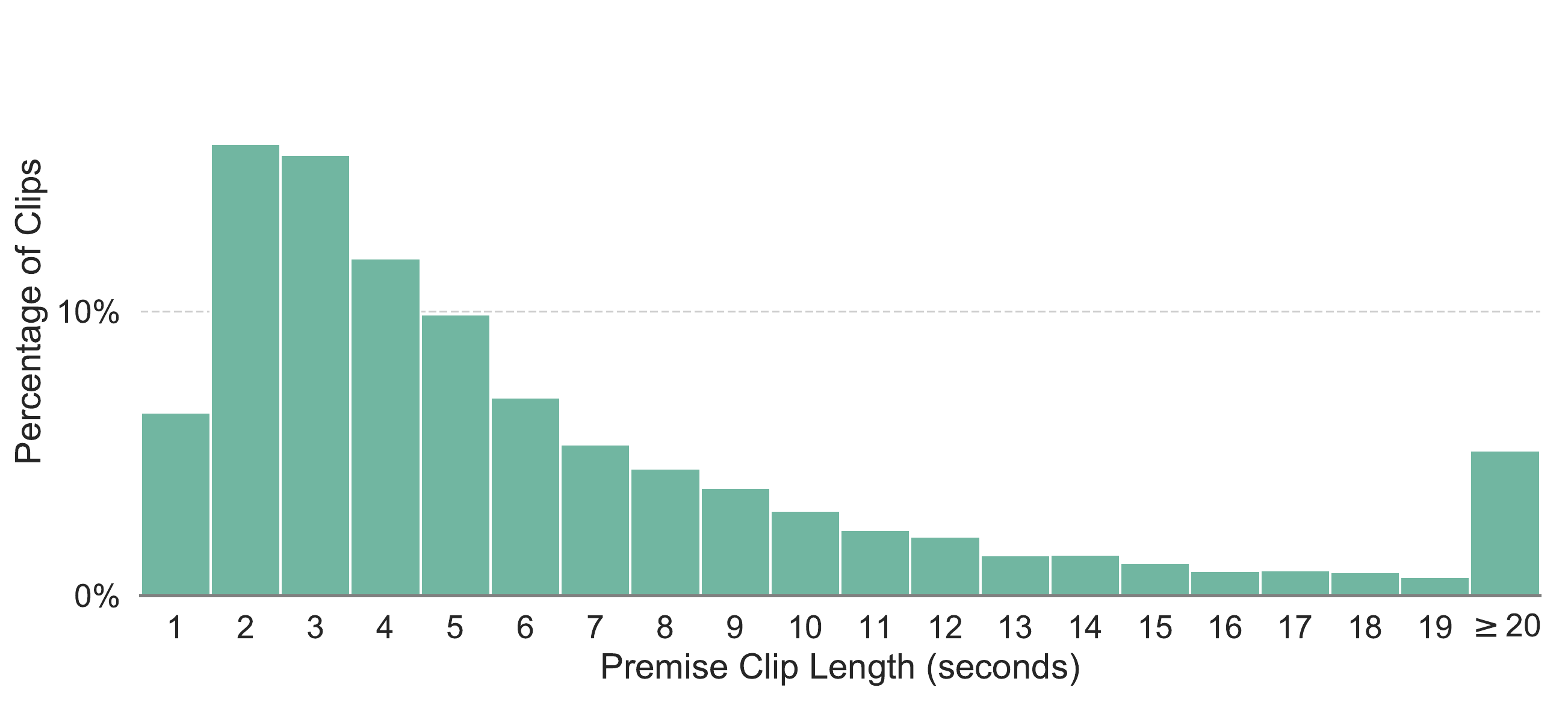}
    \vspace{-15pt}
  \caption{Distribution of premise event length.}
  \label{fig:premise_clip_len_dist}
   \vspace{-10pt}
\end{figure} 

\begin{figure}[t]
    \centering
  \includegraphics[width=\linewidth]{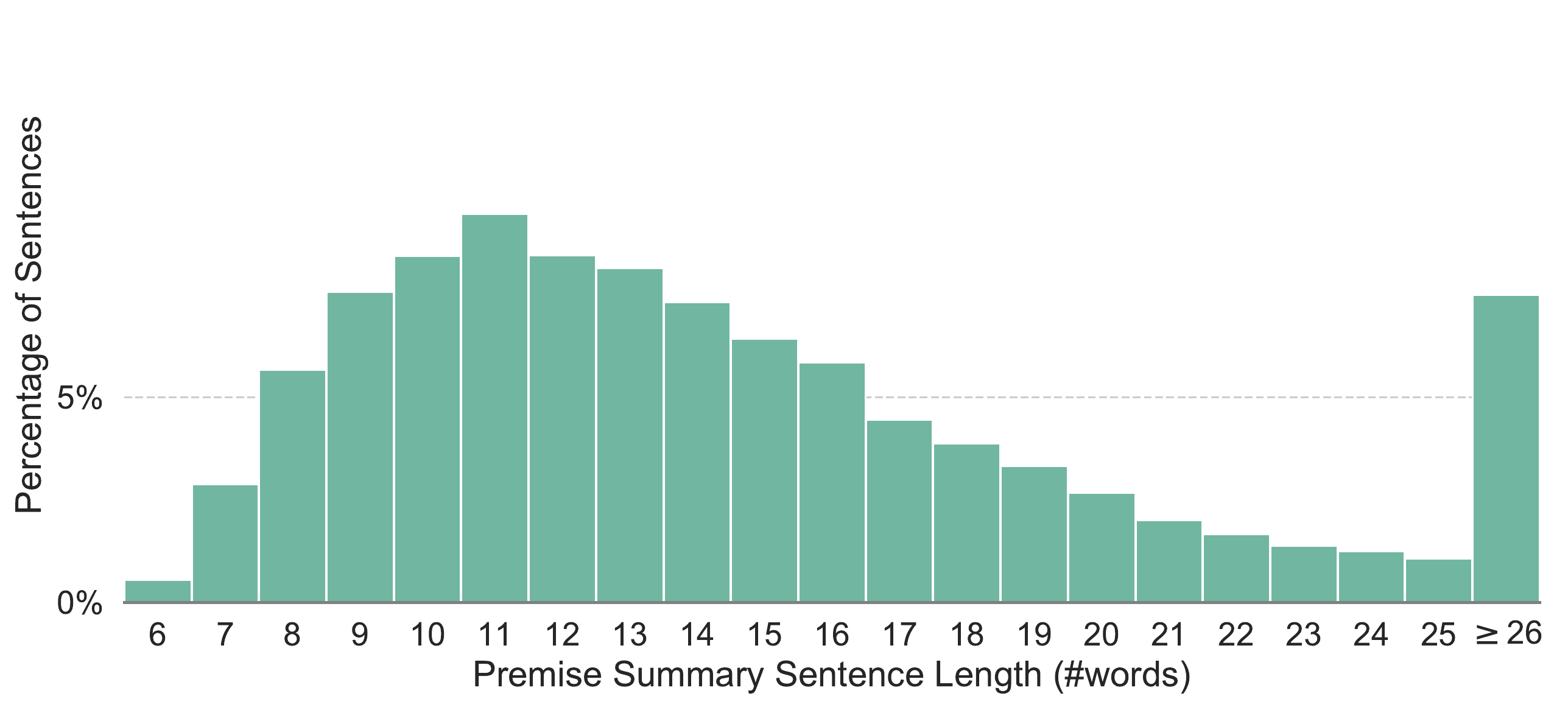}
    \vspace{-15pt}  
  \caption{Distribution of premise summary length.}
     \vspace{-10pt}
  \label{fig:premise_summary_sent_len_dist}
\end{figure}

\begin{figure}[t]
    \centering
  \includegraphics[width=\linewidth]{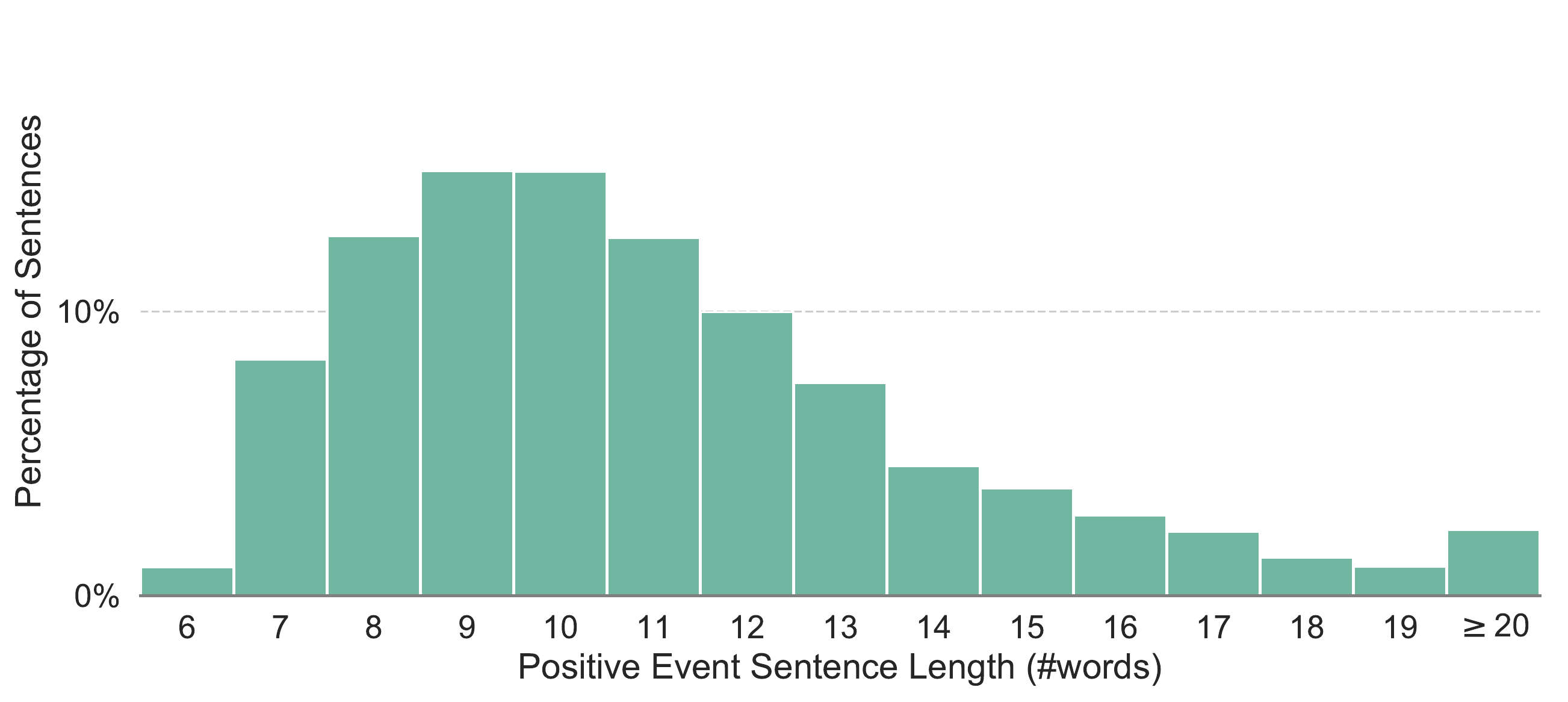}
  \caption{Distribution of positive future event length.}
  \label{fig:pos_sent_len_dist}
\end{figure} 

\begin{figure}[t]
    \centering
  \includegraphics[width=\linewidth]{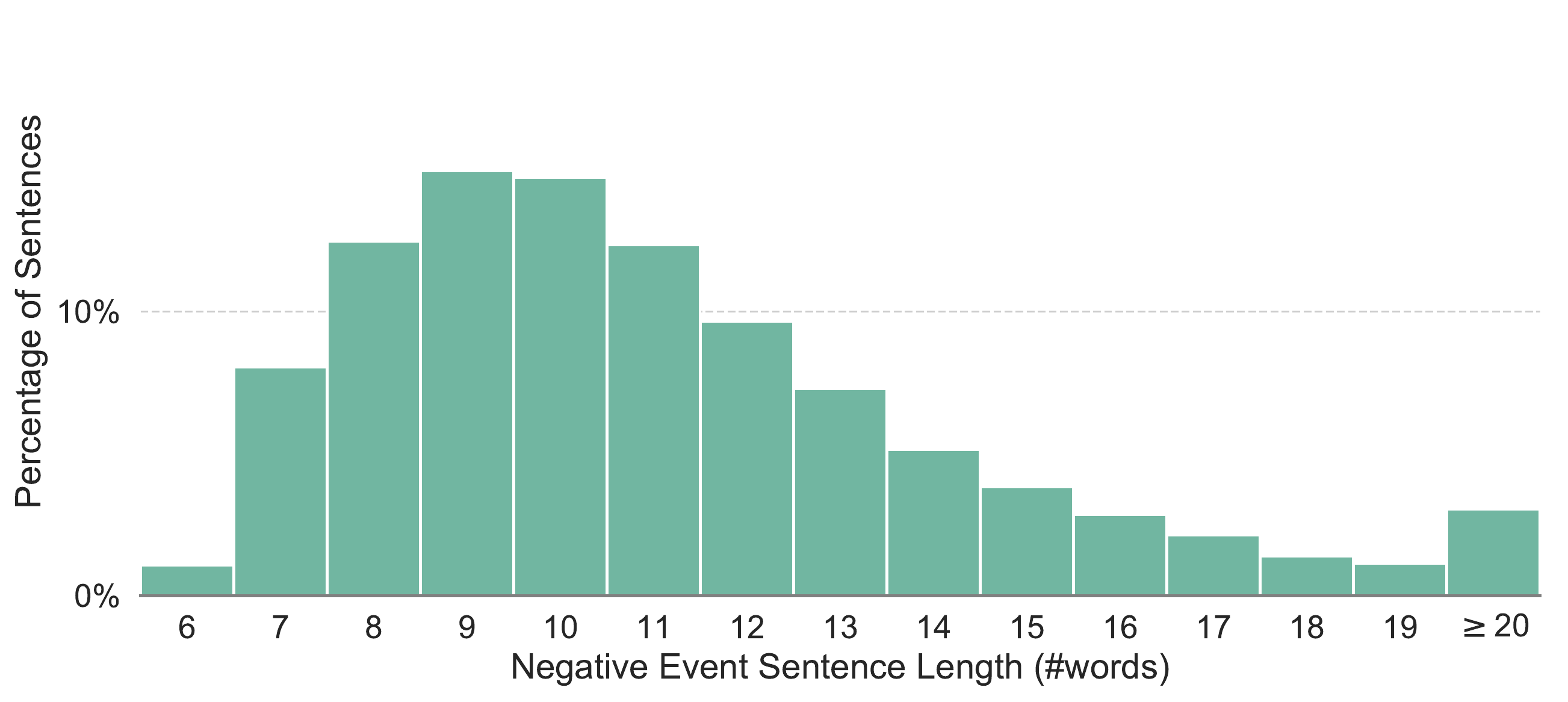}
  \caption{Distribution of negative future event length.}
  \label{fig:neg_sent_len_dist}
\end{figure}

Next, we show the distribution of premise event length and premise summary length in Figure~\ref{fig:premise_clip_len_dist} and Figure~\ref{fig:premise_summary_sent_len_dist}, respectively.
In addition, we also show the distribution of positive future event length and negative event length in Figure~\ref{fig:pos_sent_len_dist} and Figure~\ref{fig:neg_sent_len_dist}.

\subsection{Additional Data Collection Details}\label{subsec:additional_data_collection_details}
\begin{figure*}[t]
  \centering
  \includegraphics[width=\textwidth]{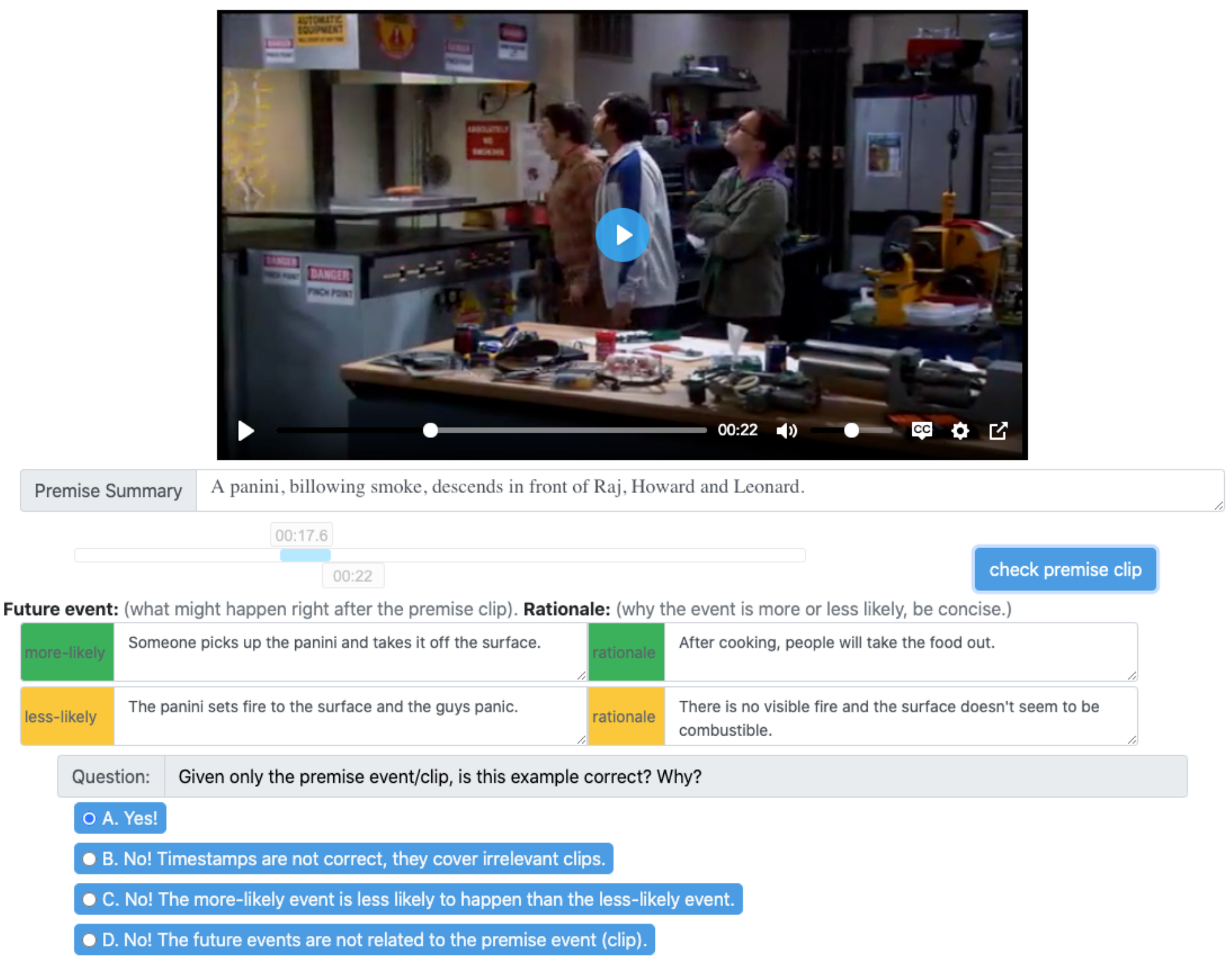}
  \caption{Example question from our qualification test. Workers have to correctly answer 7 out of 10 questions in the test to participate in our annotation task.}
  \label{fig:qual_test_example}
\end{figure*}

We hire workers from Amazon Mechanical Turk (AMT) to annotate our data. 
To ensure our data quality, we only allow workers from English-speaking countries to participate in our task. 
We require workers to have at least 500 HITs approved with an approval rate of 95\%.
Furthermore, we design a qualification test with 10 multiple-choice questions to ensure that workers well understand our annotation requirement. 
We show an example question from our qualification test in Figure~\ref{fig:qual_test_example}.
The workers have to correctly answer at least 7 questions to pass the test.
In total, 518 workers participated the test, with a pass rate of 56\%.
During data collection, we set up an automatic tool to check if all required annotations have been performed. 
We also manually review the submitted results and provide prompt feedback to them, encouraging better annotation.

\begin{figure*}[t]
  \centering
  \includegraphics[width=\textwidth]{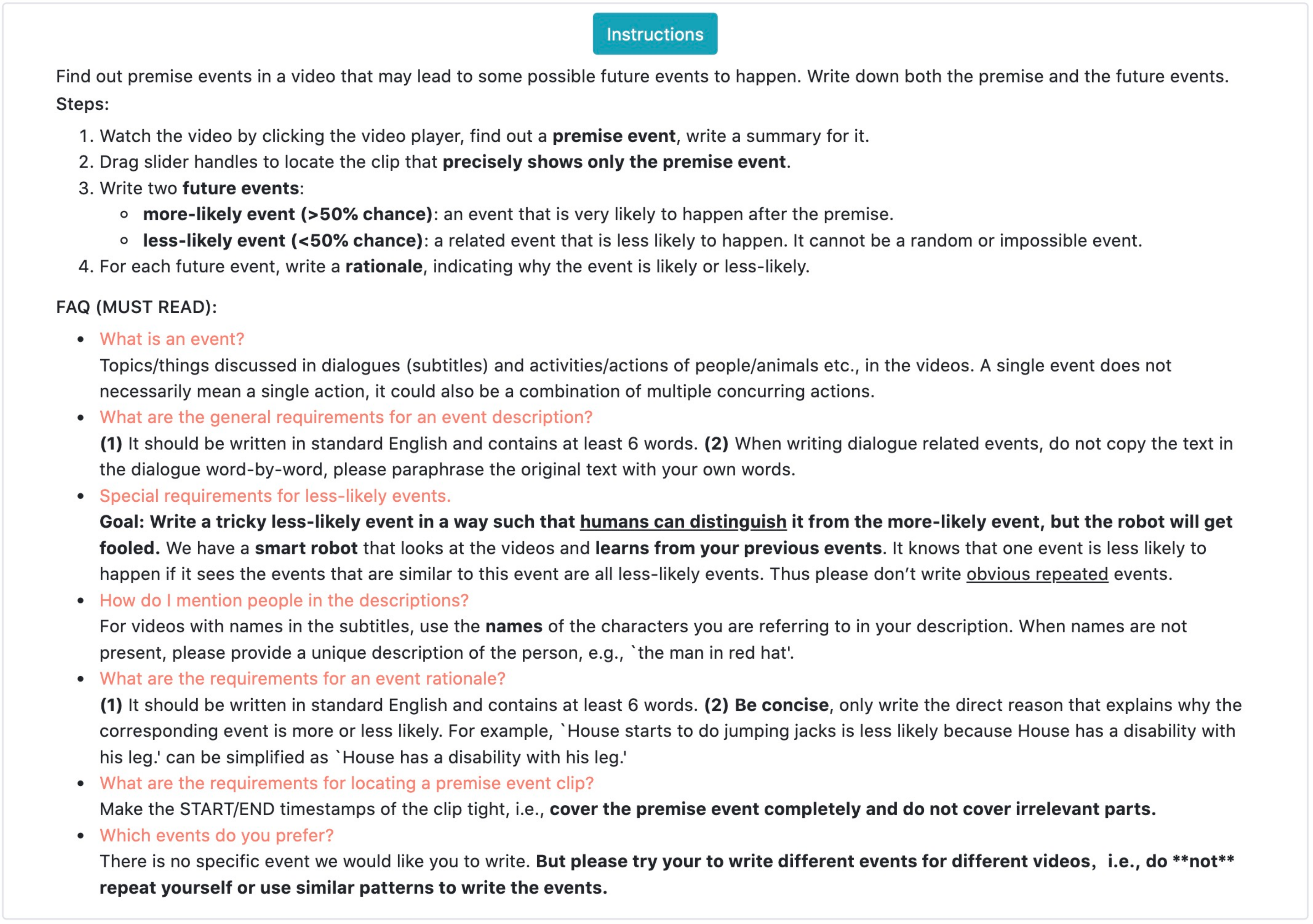}
  \caption{Annotation instructions for round two (adversarial data collection).}
  \label{fig:hit_instruction}
\end{figure*}

\begin{figure*}[t]
  \centering
  \includegraphics[width=\textwidth]{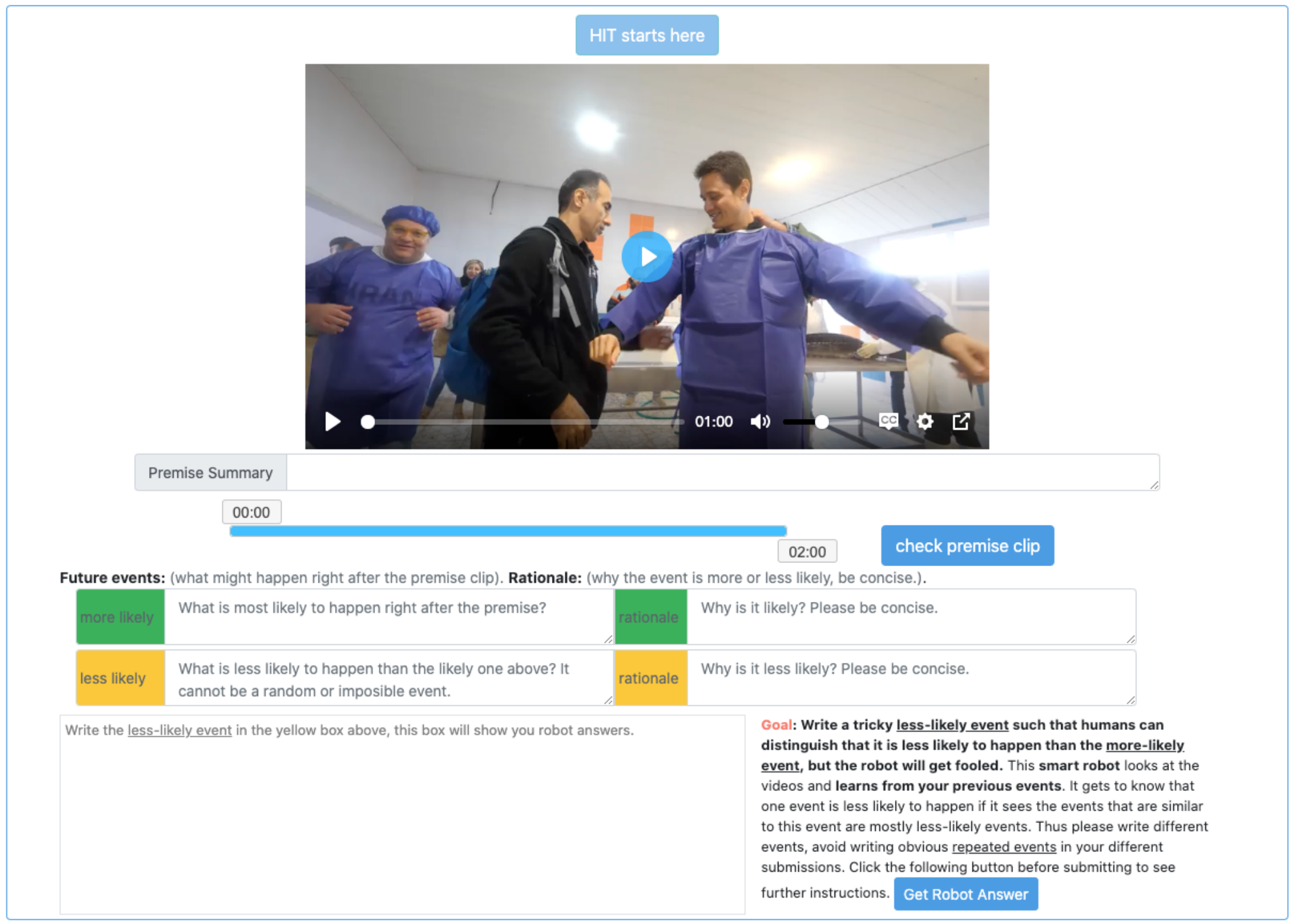}
  \caption{Annotation interface for round two (adversarial data collection).}
  \label{fig:hit_interface}
\end{figure*}

Our data collection instructions and interface for round two (adversarial data collection) are shown in Figure~\ref{fig:hit_instruction} and Figure~\ref{fig:hit_interface}, respectively. 
Round one collection details are similar to that of the round two, except that we do not require workers to fool our basic models (\textit{robot}).
In our annotation process, the actual future events in the videos are not hidden from the workers to ease the collection. The workers can either write the actual future event as the more likely event, or they can hypothesize one when the actual future event in the given video is surprising/rare (such as some events in sitcoms). 
To ensure the quality of the examples, we conduct a strict filtering step in which each example is verified by three extra workers (verifiers) and we only accept examples where at least three out of four (one writer + three verifiers) reach an agreement, as~\citet{anne2017localizing,nie2019adversarial}.

\subsection{More Results}~\label{subsec:more_results}
\paragraph{Oracle Premise Results.} As an oracle test, we apply the collected premise summary as an auxiliary input to the model, removing certain obstacles of video-dialogue understanding in our baseline model.
We show this oracle model performance in Table~\ref{tab:oracle}.
Our model with premise summary (oracle) achieves 75.64\%, which is significantly higher than the one without it (67.46\%), indicating the desire for better video-dialogue understanding.

\begin{table}[!t]
\centering
\small
\setlength{\tabcolsep}{4pt}
\scalebox{0.9}{
\begin{tabular*}{\linewidth}{l@{\extracolsep{\fill}}r}
\toprule
Model & Accuracy (\%) \\
\midrule
video + dialogue + future & 67.46 \\
\midrule
\quad + premise summary (oracle) & 75.64 \\ 
\bottomrule
\end{tabular*}
}
  \vspace{-4pt}
\caption{Oracle performance with premise summary.}
\label{tab:oracle}
\vspace{-8pt}
\end{table}

\paragraph{Future Event Generation Results.} Given the videos, we can also set up an alternative task of using a captioning-style model to generate future event descriptions. 
We use the MultiModal Transformer from~\citet{lei2020tvr} as our baseline for this task. 
This model uses a standard transformer encoder-decoder architecture for caption generation. 
The video embeddings and dialogue embeddings are concatenated as inputs~\cite{lei2020mart} to the transformer encoder.
We use the default model and training configurations from~\citet{lei2020tvr}.
With this system, we evaluate generation performance with video and dialogue as inputs. 
Our video+dialogue model has CIDEr-D~\cite{vedantam2015cider}: 19.57, BLEU@4~\cite{papineni2002bleu}: 1.80, Rouge-L~\cite{lin2004rouge}: 16.42, and METEOR~\cite{denkowski2014meteor}: 7.58. 
Note that we only use this generation task to demonstrate that it is possible to generate future event sentences from videos. 
This may not be as suitable as our default multiple choice setup to serve as an benchmark, since generation is known to be more difficult to evaluate~\cite{liu-etal-2016-evaluate}.
Besides, it also requires multiple references~\cite{vedantam2015cider} to be more accurate. 
Therefore, we recommend future work to use human evaluation if you pursue a generation-based setup on our dataset.

\subsection{More Qualitative Examples}\label{subsec:more_qualitative_examples}
\begin{figure*}[t]
  \centering
  \includegraphics[width=\textwidth]{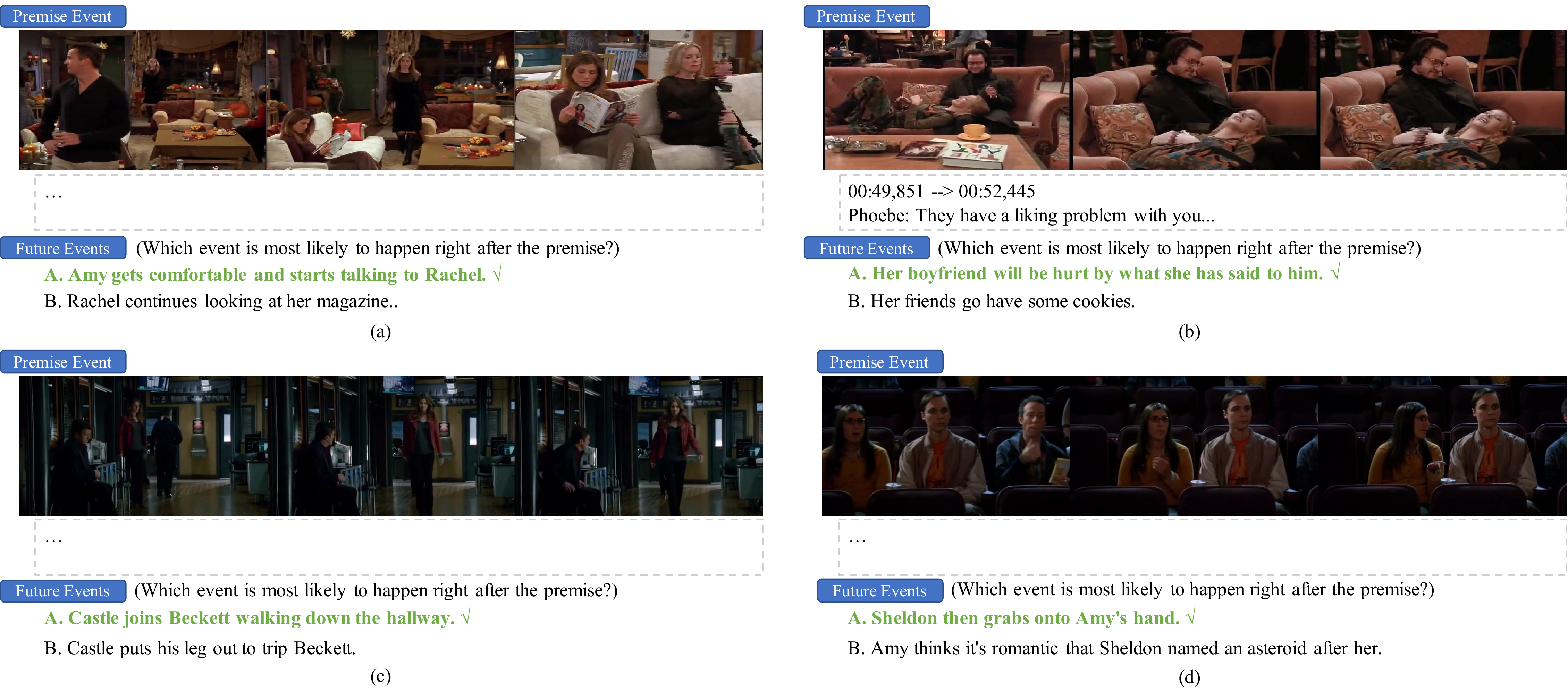}
  \caption{Correct prediction examples from our best model. Left column shows human annotated examples, right column shows adversarial matching examples. Ground truth answers are in bold and green, model predictions are indicated by \checkmark.}
  \label{fig:supp_model_correct_pred_examples}
\end{figure*}

\begin{figure*}[t]
  \centering
  \includegraphics[width=\textwidth]{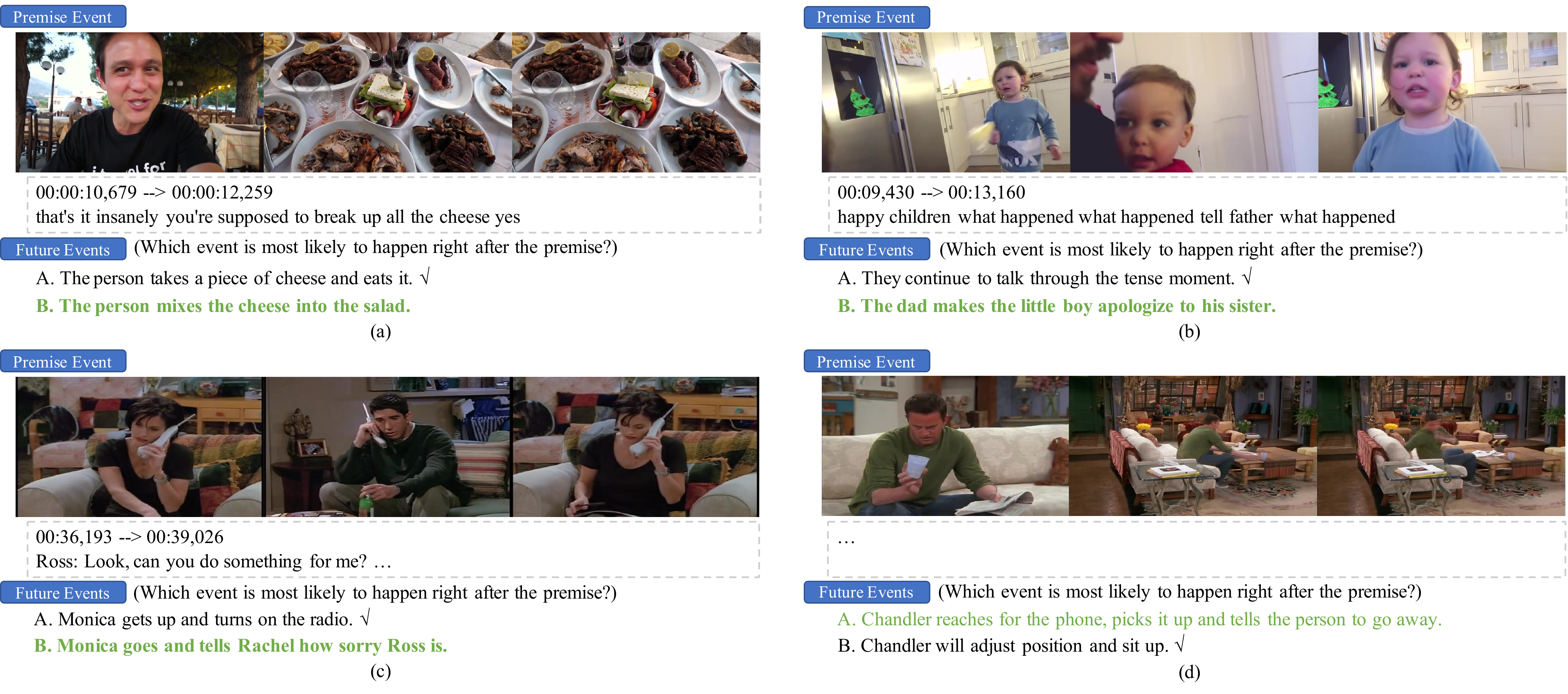}  
  \caption{Failure examples from our best model. Left column shows human annotated examples, right column shows adversarial matching examples. Ground truth answers are in bold and green, model predictions are indicated by \checkmark.}
  \label{fig:supp_model_wrong_pred_examples}
\end{figure*}

We show more correct and incorrect predictions from our best model (video + dialogue + future) in Figure~\ref{fig:supp_model_correct_pred_examples} and Figure~\ref{fig:supp_model_wrong_pred_examples}, respectively.

\clearpage

